# Out-of-distribution data supervision towards biomedical semantic segmentation

Yiquan Gao, Duohui Xu

University of Leicester, University Rd, Leicester LE1 7RH, United Kingdom

## ABSTRACT

Biomedical segmentation networks easily suffer from the unexpected misclassification between foreground and background objects when learning on limited and imperfect medical datasets. Inspired by the strong power of Out-of-Distribution (OoD) data on other visual tasks, we propose a data-centric framework, Med-OoD to address this issue by introducing OoD data supervision into fully-supervised biomedical segmentation with none of the following needs: (i) external data sources, (ii) feature regularization objectives, (iii) additional annotations. Our method can be seamlessly integrated into segmentation networks without any modification on the architectures. Extensive experiments show that Med-OoD largely prevents various segmentation networks from the pixel misclassification on medical images and achieves considerable performance improvements on Lizard dataset. We also present an emerging learning paradigm of training a medical segmentation network completely using OoD data devoid of foreground class labels, surprisingly turning out 76.1% mIoU as test result. We hope this learning paradigm will attract people to rethink the roles of OoD data. Code is made available at https://github.com/StudioYG/Med-OoD.

## 1. INTRODUCTION

Semantic labeled data is often exiguous in the medical domain as the medical imagery is hard to gather due to data protection protocols and the professional pixel labeling from pathologists or clinicians is highly costly and laborious. Therefore, many medical semantic datasets are usually limited and imperfect [53,39,38,60,15,5]. Training on these datasets tends to restrict the fully-supervised semantic segmentation network to underperform on the target visual tasks and make the segmentation network suffer from spurious correlations, thereby inducing the misclassification between foreground and background concepts.

Spurious correlation [50,35,12,59] is an unsolved essential problem in classification and semantic segmentation fields, which significantly influences the most misclassifications on unseen data samples. For instance, a resulting misclassification of medical segmentation is reflected in Figure 1 where the model erroneously confounds a large piece of foreground region (like connective tissue) to be a background concept. Copious endeavors [30,6,4,46,29] aim to address the misclassification by using Out-of-Distribution (OoD) data to learn discriminable feature representations to separate the foreground and background cues. For image classification, in-distribution (ID) data would be images containing classes of interest (e.g. dog and cat) while out-of-distribution data would be images (e.g. car and bike) other than the classes. Similarly for semantic segmentation (which is actually pixel classification), image pixels containing pre-defined foreground classes would be in-distribution while pixels outside of the foreground classes would be out-of-distribution. Therefore, in semantic segmentation ID data is images including the foreground class pixels, whereas OoD data refers to images only containing non-foreground pixels, not necessarily limited to background pixels [29,19,21]. The main concerns of current methods using OoD data are to suppress the misclassification problem of either imbalance classification [30,57] or weakly supervised semantic segmentation [29], but few attempt focuses on the misclassification of fully-supervised semantic segmentation. The success of these methods on natural images largely relies on collecting well-curated OoD images from the external data sources or multiple environments and learning based on sophisticated feature regularization objectives. However, unlike natural images, to find an appropriate external data source for building OoD dataset is extremely knotty for a medical segmentation task. The OoD collection procedure also incurs a certain amount of additional annotations. Moreover, learning with existing feature regularization objectives is prone to mistakes from domain-specific priors as medical images have different characteristics from natural images. Thence, in order to tackle the misclassification of medical segmentation (like Figure 1), our work takes inspiration from recent efforts advocating the use of OoD data, but our goal is to introduce the OoD data supervision idea to fully-supervised biomedical segmentation network to optimize the latent feature correlations in semantic segmentation and mitigate the misassignment of unrelated semantics. We explore to construct a simple yet efficacious OoD dataset diametrically from the medical task dataset without resorting to external data source or requiring any additional annotations. We investigate the impact of positive-negative sample ratio to pixel misclassification based on empirical experiments. No need of additional feature regularization objective to provide informative priors, we learn directly from the estimated amount of OoD data and the ID data to impart complementary information supervision for promoting the medical semantic segmentation in case of a limited set of ID data.

To summarize, the major contributions of our work are five-fold:

1) We propose a novel data-centric framework, namely Med-OoD of mining the OoD samples from the medical task dataset instead of external data sources, which opens a new viewpoint to refine the generalization power for fully-supervised semantic segmentation in case of the limitation of task data quantity.

2) We investigate and observe that the positive-negative sample ratio plays an important effect on the misclassifications of semantic segmentation. Ablation study shows that the range of optimal balance ratio is existing and approaching the optimal balance ratio is able to diminish misclassified pixels in semantic segmentation.

3) We introduce a numerical method to rationally estimate the percentage of OoD dataset to use in order to promote the overall positive-negative sample ratio close to the optimal balance point.

4) We discover that the addition of more OoD data into training does not mean to gain better generalization which conflicts to the common perception: more training data often bring about stronger performance. We empirically show that batch normalization is critical to exert the power of OoD samples for improving the generalization of medical semantic segmentation.

5) We reveal a promising learning paradigm to train a medical segmentation network only based on OoD data (totally free of foreground class labels), which can save much labor for annotating the dense foreground objects in ID data, and the performance is surprisingly up to 76.1% mIoU on medical test set.

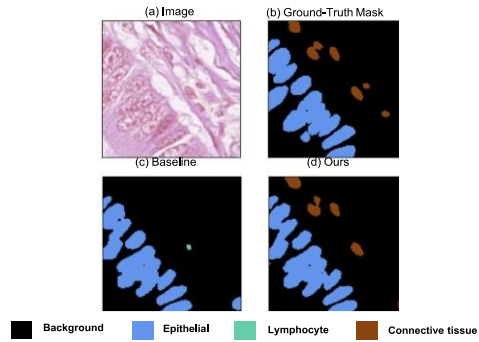

Figure 1. Misclassifications often happen: (c) Baseline confounds the foreground region of Connective tissue as background concept. (d) Our Med-OoD rectifies misclassifications on large region of Connective tissue by using OoD data as negative samples to optimize the latent feature correlations.

## 2. RELATED WORK

**Medical semantic segmentation**. A noticeable step in medical semantic segmentation challenges has been made by the seminal work of Unet [43] which employs the crop and concatenation operations to eliminate the least important edge-pixel feature information and propagate the major contextual information to greatly improve the object segmentation in a local area using the context signals of a bigger overlapping area. Another prominent variant called Unet++ [62] has taken advantage of the skip connections of DenseNet [22] as an intermediary grid to impart more useful semantic signals between the contracting and expansive paths for higher segmentation accuracy. However, unlike natural image datasets, medical image has more considerable limitations to face in the real world: 1) most public medical datasets [55,16,11] are restricted in smaller data scale compared to natural datasets [7,33]. 2) data sharing of separate medical images between different organizations through large publicly available databases are severely obstructed by the risk to patient privacy due to the concerns of imperfect data anonymization [47]. 3) annotating medical images is much more expensive compared to natural images, which requires the professional knowledge and experience of medical domain experts [54], and pixel-level annotations on medical images cost greater labor and time from the experts, hence, pixel annotations are more scanty than weaker annotations for medical images such as image-level label. These limitations significantly prevent the fully convolutional networks e.g., Unet and Unet++ from unlashing the prospective segmentation performance in a variety of realistic medical tasks towards the precision medicine.

**Learning with Out-of-Distribution data**. Data-centric frameworks represents a fresh tendency underlining that the systematic design and engineering of data is a key point to establish powerful machine learning based systems [25,48,58,61]. Out-of-Distribution (OoD) data based learning methods are a main stream of data-centric computer vision to tackle the limited learning tasks in the shortage of In-Distribution (ID) images and labels. Open-sampling method [57] demonstrates the use of OoD data to resolve the in-distribution class imbalance problem of long-tailed datasets in a Bayesian perspective which is to rebalance the class priors with open-set noisy labels and push forward the network to learn distinguishable representations. This exemplar essentially indicates that OoD data is crucial to rebalance the intrinsic data imbalance ratio unavoidable to the most training datasets. Furthermore, [8] and [20] both aim at adopting a carefully selected OoD dataset to regularize the learning of class posteriors in order to lessen the confidence of OoD data to detangle the spurious correlations between OoD and ID data. W-OoD [29] is a closely related work to ours, which extracts an OoD dataset from an external vision data source and regularizes a feature learning objective to enlarge the distance between hard OoD and ID data in the latent feature space to restrain the classifier from the misclassifications that are resulted by spurious correlations. The success of W-OoD on the weakly-supervised semantic segmentation is attributed to the diversity of feature contrasting between per ID sample and corresponding hard OoD samples since in natural images, one ID image is frequently corresponding to various hard OoD images (in high misclassified probability) with distinct background cues and the diversity of OoD samples result in vast OoD feature fluctuation in latent feature space, which benefits the feature regularization learning to reduce the spurious correlations [50,35,44,41]. However, unlike natural images, medical image exists a series of particular image characteristics [36] such as specific-range distributions of intensity and texture, making medical image more stable than the natural image [3]. Therefore, medical OoD samples have much less diversity and smaller feature fluctuation in the latent feature space so that the idea of learning separable feature representations between hard OoD and ID data through a feature learning objective is not very feasible and effective for medical image segmentation. Despite this, the behavior of introducing OoD data in W-OoD inspires us to reconsider a potential assumption that data ratio between positive and negative samples can importantly influence the misclassifications and introducing OoD dataset can remedy the original dataset to approach the optimal balance ratio to optimize the latent feature correlations in semantic segmentation and mitigate the misassignment of unrelated semantics. Besides, it is worth noting that WOoD focuses on weakly-supervised semantic segmentation on natural images using OoD data but ours concentrate on fully-supervised medical semantic segmentation using OoD data.

## 3. METHODOLOGY

We propose a data-centric framework called Med-OoD to make the fully-supervised biomedical segmentation benefit from efficient OoD data supervision. The framework sequentially describes the construction procedure (as shown in Figure 2) of OoD data in Sec. 3.1, the numerical method to estimate the suitable percentage of OoD dataset to be allocated for training in Sec. 3.2, and how to train a segmentation network using the combined OoD and ID dataset in Sec. 3.3.

### 3.1 Constructing the OoD data

How to source the unfiltered OoD data. Semantic segmentation task counts on the pixel-wise labels on a set of training images. Pixel-wise labeling process of medical task is conducted by experienced experts to determine for each pixel in an image whether the pixel contains one of the foreground classes pre-defined in a class list $C$ and if yes, assign each pixel with the foreground class labels. The set of labeled images containing the foreground classes is referred to as the ID set. While a byproduct of this labeling process is a set of unlabeled candidate images which have been determined to contain none of the foreground classes. This set of unlabeled candidate images is referred to as the unfiltered OoD set. However, in most publicly accessible medical datasets, the unlabeled byproduct of labeling process has been discarded and not included in release. Moreover, it is practically infeasible and unaffordable to acquire the unfiltered OoD set through reproducing the pixel-wise labeling process for these medical datasets. Another manner to obtain the unfiltered OoD data is to search for an analogous medical dataset which has image samples without any foreground classes designated in list $C$, but sharing similar image semantics making them likely confused by segmentation networks to be containing the target foreground classes. Therefore, this set of similar image samples can be extracted from the analogous medical dataset as the unfiltered OoD set. For example, for Lizard dataset [13] (a benchmark of nuclear segmentation), PanNuke [11] and

HuBMAP-20 [1] can be the analogous medical datasets providing bemused image samples to be used as the unfiltered OoD set. However, the disadvantage of this manner is requiring to find an external medical dataset suitable to extract the unfiltered OoD set, since for many medical datasets, to find an external dataset simultaneously meeting two conditions: 1) sharing close image semantics and 2) free from target foreground classes is arduous and even impracticable. Therefore, to ensure the universality of our method, we exploit a more cost-effective way to source the unfiltered OoD set directly from the ID data of original medical dataset which in our case is Lizard dataset. In specific, we remove all foreground-class objects (totally 6 classes) from the ID training set of Lizard using the provided pixel-level category labels to emulate the unfiltered OoD set.

Ranking and pruning to enhance OoD quality. The collected set of unfiltered OoD samples is practically imperfect since it encompasses redundant OoD candidates that are not misclassified by a segmentation network to be containing any foreground class, which means they cannot supply auxiliary negative supervision to prevent the segmentation network from confusing spurious background cues. The redundancy of OoD data can rather incur overfitting issue to the training of segmentation network. In order to improve the OoD data quality, we utilize the segmentation network, which is trained on the ID training set with foreground objects and the corresponding pixel-wise category labels, to rank the unfiltered OoD data in accordance with the predicted segmentation score $mIoU(OoD, GT)$ where $OoD$ is a OoD image sample and $GT$ is the ground truth mask of OoD image with all the values of foreground class channels as zero. When $mIoU(OoD, GT) <$ 1.0, it indicates that the OoD image is erroneously predicted by the segmentation network to have the foreground classes on its mask. We prune the unfiltered OoD set via $mIoU(OoD, GT) = 1.0$ to filter out the OoD data redundancy, consequently returning the dependable samples for the OoD dataset.

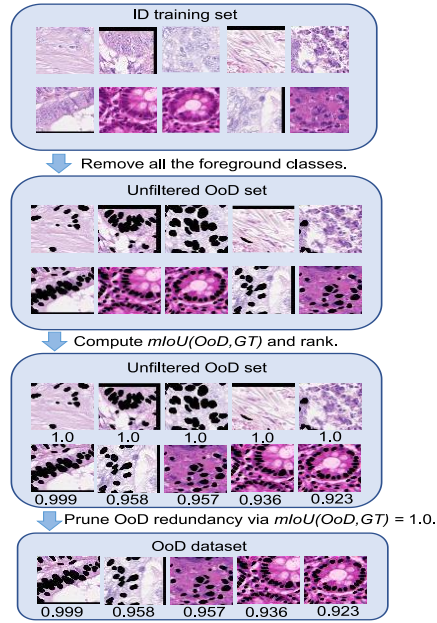

Figure 2. Constructing the OoD data: starting with the ID training set, the procedure removes each foreground class $c \in C$, prunes out OoD redundancy from unfiltered OoD set and finally returns reliable OoD dataset.

### 3.2 Estimating the OoD proportion

Directly integrating the resulting OoD dataset with the ID dataset to train the segmentation network leads to a degenerated performance as illustrated in Sec. 4.3 and Table 2 on account of the imbalance ratio between positive and negative samples. The observation on the curve in Figure 5 proves the importance of tuning the positive-negative sample ratio to the optimal balance scope. An image patch is defined as a positive sample as long as it contains any foreground pixels while it is defined as a negative sample so long as it contains any non-foreground pixels. Accordingly, an ID image patch can be considered as a positive sample and a negative sample if it contains both foreground and background pixels but an OoD image patch is just counted as a negative sample since it is exclusive of foreground pixels. However, it is unable to gauge straightforward what percentage of the OoD dataset would be rational amount to attain the optimal balance in positive-negative sample ratio for the combined dataset. We address this by proposing a numerical estimation method to determine what percentage of the OoD dataset to be allocated. First the calculation of positive-negative sample ratio for a combined dataset is expressed as:

$$pnr(D_{com}(D_{ood}, D_{id})) = \frac{|D_{com}^{pos}|}{|D_{com}^{neg}|} \quad (1)$$

where *pnr* denotes the positive-negative sample ratio of a dataset and $D_{com}(D_{ood}, D_{id})$ means the combined dataset of both out-of-distribution dataset $D_{ood}$ and in-distribution training dataset $D_{id}$. $D_{com}^{pos} \subseteq D_{com}(D_{ood}, D_{id})$ refers to the positive subset of the combined dataset and $D_{com}^{neg} \subseteq D_{com}(D_{ood}, D_{id})$ refers to the negative subset. $|*|$ denotes to get the length of a dataset. Subsequently we devise a numerical method to estimate the rational amount of the OoD data by minimizing an objective, as follows:

$$pct_{opt} = \arg\min_{pct \in \Theta} f_T(pct) \quad (2)$$

where *pct* means the percentage of $D_{ood}$ and $\Theta$ is a pre-defined percentage search space which is {0,...,1} with the interval of 0.1. $pct_{opt}$ denotes the optimal percentage of $D_{ood}$ estimated by finding the best *pct* value from the search space $\Theta$ to minimize the $f_T(pct)$ while $f_T(pct)$ is the numerical optimization objective defined as the absolute difference between two terms:

$$f_T(pct) = |pnr(D_{com}(pct \cdot D_{ood}, D_{id})) - pnr_{opt}| \quad (3)$$

where one is the concrete *pnr* calculated on combined dataset and another is the optimal balance reference point $pnr_{opt}$ for positive-negative sample ratio, which in the experiments we set to 0.65 as explained in Sec. 4.3 and Figure 5. $pct \cdot D_{ood}$ refers to the random selection of *pct*(%) samples from $D_{ood}$. The purpose of $pct_{opt}$ is to allot the rational percentage of $D_{ood}$ to push the whole positive-negative sample ratio of combined dataset towards the optimal balance reference point as close as possible, which enhances the mutual information and ensures the well-balanced gradient optimization between ID and OoD samples when training a segmentation network.

### 3.3 Learning with combined OoD and id dataset

The combined dataset formed by the complete OoD dataset $D_{ood}$ and ID training dataset $D_{id}$ is first imported into the Eq. 2 for finding the best solution towards the objective. The optimal percentage $pct_{opt}$ estimated by Eq. 2 is used to randomly select out $pct_{opt}$ samples from $D_{ood}$ as the final OoD dataset $D_{ood}^{opt}$. $D_{ood}^{opt}$ is then integrated with $D_{id}$ as a combined dataset $D_{com}(D_{ood}^{opt}, D_{id})$. OoD samples use the zero-pixel masks and ID samples use the prepared pixel-wise category masks as their ground-truths. The segmentation loss for training our segmentation networks is then shown below:

$$L_{seg} = \frac{1}{|C|} \sum_{c=1}^{|C|} [L_{Dice}(F_c(I_{id}), GT_c) + \lambda \cdot L_{Dice}(F_c(I_{ood}), 0)] \quad (4)$$

where $F_c$ is the predicted segmentation mask for foreground category *c*, we adopt the Dice loss [51] to learn against each foreground-class ground-truth mask $GT_c$ for in-distribution images $I_{id}$ and against zero ground-truth mask for out-of-distribution images $I_{ood}$. $\lambda$ is a coefficient of weighting the loss term of $I_{ood}$. We apply the shuffle operation to $D_{com}(D_{ood}^{opt}, D_{id})$ to ensure the OoD and ID samples are assigned to each batch with unbiased probability. The eventual $D_{com}(D_{ood}^{opt}, D_{id})$ is leveraged for training segmentation networks.

## 4. EXPERIMENTS

### 4.1 Experimental setting

In-Distribution and Out-of-Distribution dataset. Our experiments use Lizard dataset [13] as the ID dataset. It is a benchmark of nuclei segmentation with 6 foreground classes annotated in the ID images, including Eosinophil, Epithelial, Lymphocyte, Plasma, Neutrophil and Connective tissue. ID set totally provides 291 image regions with an average size of 1,016×917 pixels from various patients. We crop each image region with zero padding into small patches of 128×128 pixels, resulting in 18,374 image patches. We follow the experimental setup of [13] to split the image patches into 3 folds for cross validation and the final evaluation of each method is reported using the average test results across the 3 folds. The ID training set is denoted as $D_{id}$ in the experiments. In each fold, the ID training set $D_{id}$ follows the procedure in Sec.

3.1 and Figure 2 to construct OoD set $D_{ood}$. Through the numerical estimation method in Sec. 3.2, we obtain the final OoD dataset $D_{ood}^{opt}$ by randomly selecting out $pct_{opt}$ samples from $D_{ood}$. Following Sec. 3.3, $D_{ood}^{opt}$ is combined with $D_{id}$ as $D_{com}(D_{ood}^{opt}, D_{id})$ for subsequent processing and training segmentation networks.

Implementation details. Our Med-OoD framework can be seamlessly integrated into any medical segmentation methods. For the medical segmentation experiments, we employ state-of-the-art architectures including Unet [43], Unet++ [62] and MAnet [9]. For the backbones, Unet adopts VGGs [49] with batch normalization [24]. Unet++ takes ResNets [18] and DenseNets [22]. MAnet uses efficient MobileOnes [56], MobileNets [45] and TinyNets [17] to reduce computational burden. For fair experiments, all the backbones are initialized with pre-trained weights on ImageNet [7], adhering to the common practices [10,23,14,28,32]. The hyperparameter settings follow the original works of architecture and backbone networks. We denote each method by Architecture-Backbon and conduct two experiments per method where one is the baseline version trained on $D_{id}$ and another is the corresponding Med-OoD version trained on $D_{com}(D_{ood}^{opt}, D_{id})$. To evaluate the test results, we use two widely used metrics in medical semantic segmentation [52,37] including mean Intersection over Union (mIoU) [42] and Dice Coefficient (DSC) [34]. Regarding the hyperparameters defined in Med-OoD, we set the coefficient $\lambda = 1.0$ and the optimal balance reference for positive-negative sample ratio $pnr_{opt} = 0.65$ because the ablation study shows the scope of optimal balance ratio is (0.61,0.68) and Unet-VGG11BN reaches the mIoU peak at the ratio of 0.644, which is approximated as 0.65.

## 4.2 Result comparisons

Qualitative analysis on segmentation results. Our Med-OoD method outperforms its baseline as shown in Figure 3 by producing more precise segmentation masks, much closer to the ground truths in terms of the mask quality. In particular, the baseline presents serious misclassifications for the highly frequent foreground classes like connective tissue and epithelial in the nuclei images. Applying Med-OoD rescues the misclassifications of considerable foreground classes by the baseline to a large extent. But compared with the baseline, Med-OoD seems to have no much improvement on the segmentation of plasma, neutrophil and eosinophil, making them likely misclassified as other foreground concepts. This is attributed to the low distributions of plasma, neutrophil and eosinophil in the dataset [13], which fails to supply enough target samples for discriminating these rare foreground classes from other foreground classes.

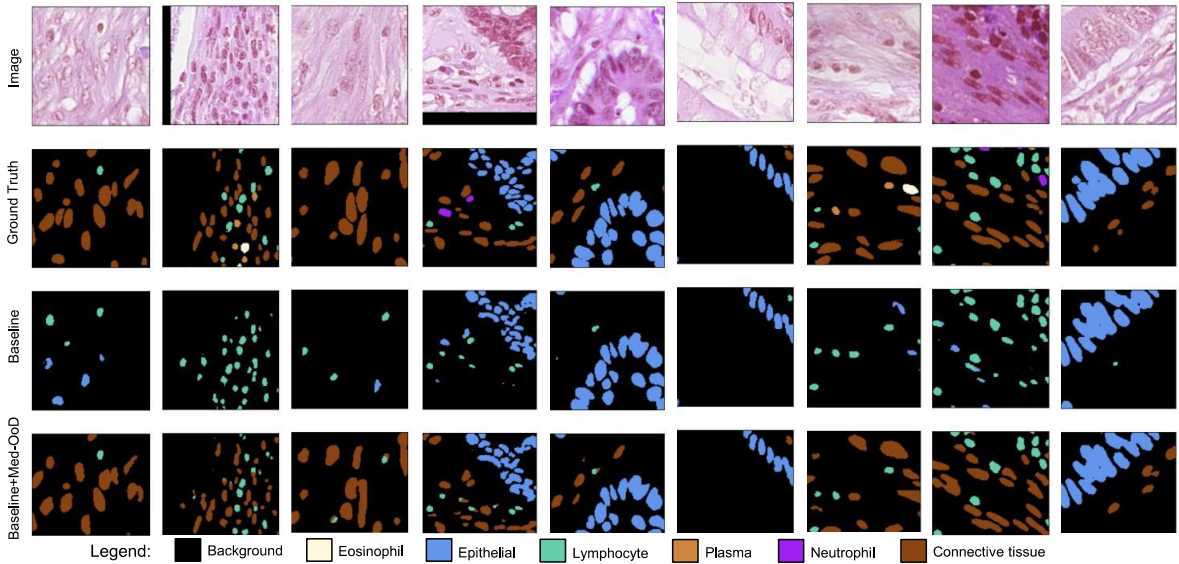

Figure 3. Comparison of segmentation results on Lizard test set: from top to bottom it respectively shows image (first row), ground truth (second row), segmentation mask of baseline (third row) and segmentation mask of baseline+Med-OoD (last row). Here the baseline is Unet-VGG11BN.

Quantitative analysis on segmentation results. Our method can be simply applied to other semantic segmentation methods without any modification on the original baseline, since it is a purely data-centric framework. We employ our method to multiple prevalent medical segmentation methods. Table 1 illustrates the segmentation results for each corresponding baseline, followed by their respective performances after utilizing Med-OoD framework. We can see that Med-OoD typically improves the both metrics (i.e., mIoU and DSC) by a substantial margin for all the medical segmentation methods, which indicates that the quantity of misclassified pixel regions are largely subdued and decreased by Med-OoD and therefore, the refined segmentation masks exhibit more meticulous foreground patterns highly approaching the semantic ground truths as viewed in Figure 3. The observations on $\Delta pnr$ signify that the numerical estimation method of Med-OoD is able to push the positive-negative sample ratio (*pnr*) of training set closer to the $pnr_{opt}$ by minimizing the objective in Eq.3 and rationally integrating OoD samples into training. Notably, Med-OoD remarkably increases the mIoU value by 1.6% for the first baseline, which is considered as an impressive performance gain as it is brought by merely the OoD samples mined from the original dataset without any support from external data sources or additional annotations. The performance improvements on all the methods simultaneously prove that Med-OoD is beneficial to enhancing the data efficiency on the original dataset.

Table 1. mIoU and DSC evaluations on Lizard test set: we evaluate the segmentation performance of each method and the corresponding version using Med-OoD. Med-OoD improves the results of all the methods. $\Delta pnr$ denotes the absolute distance between $pnr_{opt}$ and the positive-negative sample ratio (*pnr*) of the corresponding training set.

| Method | mIoU(%) | DSC(%) | $\Delta pnr$ |
|---|---|---|---|
| Unet-VGG11BN | 87.4 | 93.1 | 0.317 |
| +Med-OoD | **89.0** | **94.0** | **0.006** |
| Unet-VGG16BN | 88.5 | 93.7 | 0.317 |
| +Med-OoD | **89.2** | **94.2** | **0.012** |
| Unet-VGG19BN | 87.6 | 93.2 | 0.317 |
| +Med-OoD | **87.8** | **93.3** | **0.004** |
| Unet++-ResNet18 | 88.2 | 93.6 | 0.317 |
| +Med-OoD | **88.9** | **94.0** | **0.030** |
| Unet++-ResNet50d | 89.3 | 94.2 | 0.317 |
| +Med-OoD | **89.4** | **94.3** | **0.019** |
| Unet++-DenseNet169 | 88.5 | 93.7 | 0.317 |
| +Med-OoD | **89.2** | **94.2** | **0.014** |
| MAnet-MobileOneS0 | 85.5 | 91.9 | 0.317 |
| +Med-OoD | **85.9** | **92.2** | **0.019** |
| MAnet-MobileNetV2 | 84.5 | 91.3 | 0.317 |
| +Med-OoD | **84.6** | **91.4** | **0.003** |
| MAnet-TinyNetd | 84.1 | 91.0 | 0.317 |
| +Med-OoD | **84.2** | **91.1** | **0.000** |

Quality of per-class segmentation. Classes of various objects suffer from different magnitudes of misclassifications predicted by the baseline. Sorted by the descending order of misclassification severity, the foreground classes are ranked as Connective tissue, Lymphocyte, Plasma, Epithelial, Neutrophil and Eosinophil. Figure 4 compares class-wise segmentation performances between the baseline and its Med-OoD version. We observe that the classes which have been improved most by our method respectively are Connective tissue (+19.6% mIoU), Epithelial (+4.0% mIoU) and Lymphocyte (+2.8% mIoU). The vast improvements on three classes significantly show the superior power of Med-OoD to redress the misclassified pixels between foreground and background objects. However, the mIoUs for another three classes remain exactly the same as their baselines. The resulting inconsistency is actually affected by the imbalanced class distributions. It is reported in Lizard [13] that the distributions of Connective tissue, Epithelial and Lymphocyte exceedingly surpass that of Plasma, Neutrophil and Eosinophil, which leads to over-optimization [31] on the former three classes than the latter three ones during the training. This imbalanced class optimization fundamentally restricts Med-OoD to improve the performance of Plasma, Neutrophil and Eosinophil. The role of Med-OoD is equivalent to establishing OoD semantic information priors to separate the foreground and background objects by providing rational amount of OoD samples to balance the gradient optimization between positive and negative regions when learning a segmentation network, but it is unable to supply meaningful supervision signals to address the inter-class misclassifications caused by the quantity gap among different classes. Therefore, for the classes including Plasma, Neutrophil and Eosinophil, Med-OoD solely facilitates to distinguish them from background objects but cannot prevent them from being misclassified as other different classes.

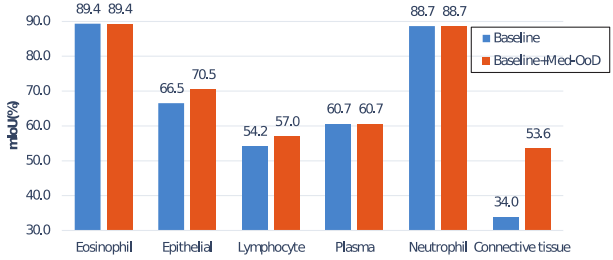
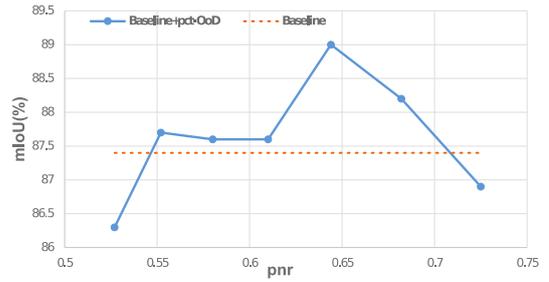

Figure 4. Per-class segmentation results: we evaluate the performances of baseline version and the Med-OoD augmented version across all the foreground classes on Lizard test set. Here the baseline is Unet-VGG11BN.

Figure 5. Adjust the percentage (*pct*) of OoD from 100% to 40% for observing the performance trend influenced by positive-negative sample ratio (*pnr*).

### 4.3 Ablation study

We conduct comprehensive ablation experiments on Lizard to empirically study the behaviors of our method. Unless specified, we use Unet-VGG11BN as the default baseline for implementing ablation experiments.

Table 2. Comparison of training size and performance among three settings by integrating different percentages *pct*(%) (e.g., 100% and 60%) of OoD data with ID data for training.

| Setting | *pct*(%) | Training size | mIoU(%) |
|---|---|---|---|
| Baseline | 0 | 12249 | 87.4 |
| $+D_{ood}$ | 100 | 21933 | 86.3 |
| $+D_{ood}^{opt}$ | 60 | 18059 | 89.0 |

Table 3. Comparison of different cases with and without batch normalization: both cases use Med-OoD for training and outclass the baseline performance on Lizard test set (i.e., 87.4% mIoU). Batch normalization has beneficial effect on Med-OoD.

| case | mIoU(%) |
|---|---|
| w/ batch-norm | 89.0 |
| w/o batch-norm | 88.2 |

**Impact of positive and negative ratio.** The segmentation quality is sensitive to the positive-negative sample ratio (*pnr*) of training data. We investigate the impact of *pnr* to the segmentation performance on test set by tuning the percentage (*pct*) of OoD data to integrate with ID data into training, as shown in Figure 5. *pnr* is increased from 0.527 to 0.725 as *pct* is decreased from 100% to 40% with an interval of 10%. It is seen that the segmentation performance attains the mIoU peak when *pnr* is 0.644 and the optimal value of *pnr* is expected to exist in the range of (0.61,0.68). Therefore, in all the experiments, we approximately set the optimal balance reference $pnr_{opt}$ of the numerical estimation method to 0.65.

On the other hand, we empirically find that increasing the amount of OoD data for the training is not meant to gain more powerful generalization. Table 2 shows the baseline (only trained on ID data) and last setting (trained on ID data and 60% of OoD data) significantly outperform the second setting (trained on ID data and 100% of OoD data) by respectively using 9684 and 3874 fewer training images than the second setting. From the results in table 2 and the performance trend curve in Figure 5, we can notice that adding more training data can boost the performance of baseline, but beyond a certain point, it will not improve model performance and even lead to an unexpected degeneration of performance, which agrees with the discovery of scaling laws [26,40,2,27] to some extent. This finding challenges against the common belief that adding more training data generally leads to stronger model performance.

**Effect of batch normalization.** Table 3 compares two case settings with and without batch normalization (BN). Two cases both use Med-OoD for training. Their architectures and all hyper-parameters are kept exactly the same. The only change is with or without BN. For the second case, we remove all BN layers in the model. We observe that the one with BN is superior to the counterpart without BN, therefore proving the beneficial effect of batch normalization to further unleash the power of OoD samples for improving the generalization. This effect probably is attributed to the role of batch normalization that decreases internal covariate shift by fixing the means and variances of layer inputs for normalizing ID and OoD data in each training mini-batch [24].

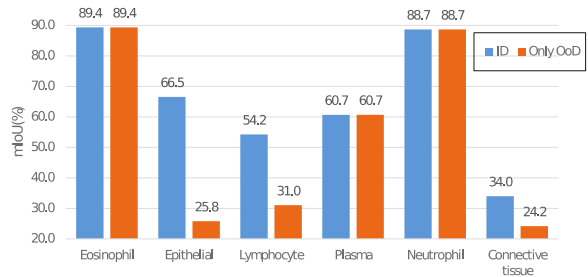

Figure 6. Foreground-class evaluation on Lizard test set. The baseline trained on $D_{id}$ and the Unet-VGG11BN trained on $D_{ood}^{opt}$ are separately denoted as ID and Only OoD.

**Effectiveness of learning OoD data.** We particularly train the Unet-VGG11BN with only the OoD training set $D_{ood}^{opt}$ instead of the combined training set $D_{com}(D_{ood}^{opt}, D_{id})$ adopted by the Med-OoD. This ensures the network training is fully on OoD data without involving any class labels of foreground objects in ID data. As shown in Figure 6, the $D_{ood}^{opt}$ trained version achieves identical mIoUs to the baseline across Eosinophil, Plasma and Neutrophil classes, even though the mIoUs for another three classes are lower than the baseline. Besides, the final evaluation (76.1% mIoU) on Lizard test set is more than expectation for the $D_{ood}^{opt}$ trained version as seen in Table 4 since it is only trained on OoD data using none of foreground-class annotations provided in ID data. Therefore, these results demonstrate the effectiveness of learning OoD data and suggest that training segmentation networks purely on OoD data can be a promising paradigm that saves much effort of labeling foreground objects in ID data and still obtains applicable performance for specific task scenes.

Table 4. Number of different foreground class samples present respectively in $D_{id}$ and $D_{ood}^{opt}$. The last column mIoU (%) denotes the final evaluation on Lizard test set for two training settings.

| Train set | Eosinophil | Epithelial | Lymphocyte | Plasma | Neutrophil | Connective tissue | mIoU(%) |
|---|---|---|---|---|---|---|---|
| $D_{id}$ | 1118 | 8739 | 7745 | 4213 | 1151 | 8688 | 87.4 |
| $D_{ood}^{opt}$ | 0 | 0 | 0 | 0 | 0 | 0 | 76.1 |

## 5. CONCLUSION

We propose a data-centric framework to provide the OoD data supervision for diminishing the misclassifications learned by medical semantic segmentation methods. With the rationally estimated amount of OoD images as negative samples complementing to the original ID images, our method is able to learn a segmentation network with more precise segmentation masks than the existing baseline methods. We empirically explore a new learning paradigm to reach adequate performance for limited task scenes by training a segmentation network purely using OoD data, which hopefully will be popularized into object detection and instance segmentation tasks.